\documentclass{article}
\usepackage{spconf,amsmath,graphicx}
\usepackage{booktabs}

\usepackage{enumitem}
\setlist{nosep, leftmargin=14pt}

\usepackage{mwe} 

\title{Runtime Freezing:\\
Dynamic Class Loss for multi-organ 3D  Segmentation}

\name{James Willoughby$^1$ ~~~~~~ Irina Voiculescu$^1$}
\address{$^1$Oxford University Department of Computer Science}
\newcommand{\rone}{$\mathit{TCF}$}
\newcommand{\rtwo}{$\mathit{PCF}$}
\newcommand{\rthree}{$\mathit{CBS}$}

\begin{document}
%
\maketitle

\begin{abstract}


Segmentation has become a crucial pre-processing step to many refined downstream tasks, and particularly so in the medical domain. Even with recent improvements in segmentation models, many segmentation tasks  remain difficult. When  multiple organs are segmented simultaneously, difficulties are due not only to the limited availability of labelled data, but also to class imbalance. In this work we propose dynamic class-based loss strategies to mitigate the effects of highly imbalanced training data. We show how our approach improves segmentation performance on a challenging Multi-Class 3D Abdominal Organ dataset.

\end{abstract}

\section{Introduction}

With the advent of isometric volumes of 3D scan data, 3D methods for automated medical image analysis are becoming increasingly common. Analysing 3D scan data is also key to crucial clinical tasks such as assessing tumor volume and borders.
In much the same way that clinicians view slices with the understanding that they are part of a known complex 3D structure, a 3D segmentation network is able to make predictions more effectively by relying on the information encoded across several slices. This is most obvious in the case of organs that appear discontinuous in some 2D slices but continuous in 3D (the Pancreas in abdominal organ segmentation).

The processing of large volumes of 3D data exacerbates a problem common to automated medical image analysis: class (or organ) imbalance. Within the same volume, some organs (liver) occupy the majority of voxels, with other organs (gallbladder) being significantly smaller.
Most multi-class medical image segmentation tasks present a significant imbalance in the class distributions. Whether this is the case of a multi-disease scenario or identifying multiple anatomical structures it is generally true that a few classes will be very common or represent most of the labelled voxels and the rest will be comparatively rare. 

In this work we propose an improvement to 3D segmentation techniques using a strategy for adjusting per-class loss dynamically. We assess its performance compared to a basic multi-class Dice loss in three distinct 3D segmentation algorithms. As part of this work we also generalise to 3D the `lambda layer' method~\cite{lambdanet}, and assess its potential.

\begin{figure}
    \centering
    \includegraphics[scale=0.7]{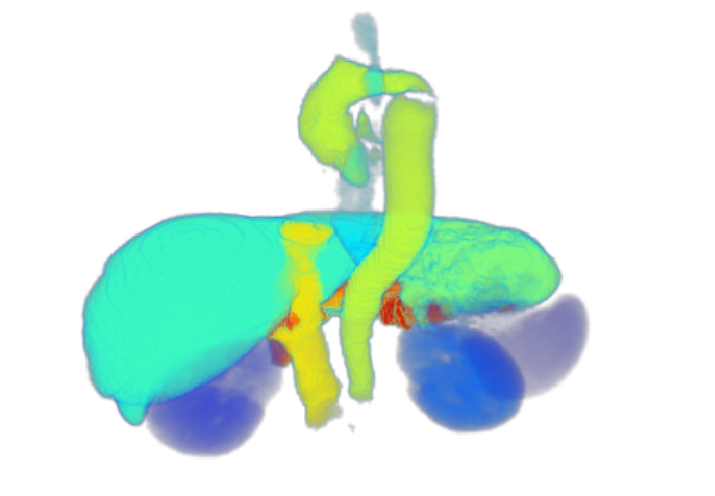}
    \caption{Abdominal 3D Segmentation Output.}
    \label{fig:enter-label}
\end{figure}

\section{Methods}

\subsection{Data}

The data used for assessing 3D segmentation performance has been a publicly available set of abdominal CT scans from the MICCAI 2015 challenge~\cite{Synapse}. This contains 120 different scans, of about 80 slices each, within which thirteen different organs are labelled.  An additional dataset has been the Totalsegmentator dataset~\cite{Totalsegmentator}, a large dataset of 1204 CT scans containing 104 anatomical structures, from which we took any CT scans which contained segmentations for six or more of the thirteen organs segmented in the challenge dataset, totalling 923 scans of 80 slices each. 

\subsection{Network Structures}
\label{sec-nets}

The 3DUNet \cite{unet3d} is a simple extension to 3D of the basic UNet concept:  the accepted input dimensions are expanded into the convolutional layers whilst keeping the main structure the same.
%
%
The UNETR \cite{hatamizadeh2021unetr} concept takes the basic structure of the UNet and replaces the convolutional blocks with blocks more similar to the Vision Transformer (ViT) ~\cite{ViT}. In ViT the input image is divided into smaller `patches' which are projected into an embedding space, mimicking the nodes in the original Transformer concept used in natural language processing~\cite{Attention}. Self-Attention is then used to construct relations between all the nodes, which is advantageous in image processing as it allows for long range interactions between sections of the image to be captured better than in other methods. The UNETR network uses a three dimensional patch, dividing the input up into large voxels instead to allow for the processing of 3D data types such as MRI and CT scans.

\subsubsection{Lambda3D}

The Lambda UNet network is similar to the UNETR except it replaces the transformer blocks with `lambda layers'~\cite{lambdanet}. These layers are also inspired by the Transformer concept except they replace the self-attention step with a more inexpensive linear attention function, which allows for the network to be trained on smaller patch sizes. This results in similar overall runtimes while still maintaining the Transformer’s ability to capture long-range interactions well. The Lambda network has been tested on 2D data and in a 2.5D capacity~\cite{ou2021lambdaunet}. 

To our knowledge, there has been no literature on a 3D extension. To add to our collection of 3D networks for the proposed dynamic scheduling experiments, we have made the additional novel contribution of extending the existing 2D Lambda network concept to full 3D. Although fine tuning it has been of no immediate interest for this particular work, its performance as a 3D segmentation network is loosely compared to the UNETR and 3DUNet.

The original Lambda network  mirrors the functionality of a Transformer, but with improved computational efficiency via a linear attention function. Self-attention for 3D is even more computationally expensive than for 2D, due to the exponential increase in patch number. However self-attention is able to capture the complex relationships of the 3D patches well. Including this novel Lambda3D network in the experiments is an opportunity to analyse how well its basic version performs in 3D segmentation in comparison to other 3D segmentation networks. Since it responds well to the proposed strategies, its potential can then be exploited in further work.

\subsection{Loss Scheduling Strategies}

Multiclass medical imaging datasets tend to be considerably imbalanced in terms of the class distributions. The class imbalance in the Multi-organ Abdominal CT Dataset is clearly shown in the leftmost column of Table~\ref{results_table}. On average, the liver represents 55.2$\%$ of the labelled voxels whereas the adrenal glands each only represent 0.1$\%$ of labelled voxels. We want our network to perform as well as possible on all classes but some classes are dramatically harder for the network to learn than others simply due to the quantity of available training data.

The loss function in a multiclass problem usually calculates a mean or sum of individual class losses. Most methods to address the impact of class imbalance function by weighting the class components of the loss function. In our case the ubiquitous Dice loss function is used throughout:
\begin{equation}
DiceLoss=1 - {\frac{2\,\mathit{TP}}{2\,\mathit{TP}+\mathit{FP}+\mathit{FN}}}    
\end{equation}
Weighting the class losses equally promotes a training pathway which works well for the more prevalent classes and not necessarily for the less prevalent ones. This is because the loss contributions for the more prevalent classes will decrease far more quickly. Weighting the loss contributions based on class prevalence, we will remove this tendency, as the loss will remain high after the prevalent classes have improved. This approach improves the network's learning for less prevalent classes. However, there are a few flaws. Firstly it can create an incredibly unstable loss function in very unbalanced cases as the loss contribution for a very small class will be weighted very highly. If the network labels even one instance of this class correctly the loss function will be hugely impacted, which makes optimisation difficult as the loss is unstable. 

Another issue with this method is that we make the assumption that the difficulty of learning a class is proportional to the quantity of class data in the dataset. This assumption ignores a number of factors which can make classes easier or harder to learn independent of class prevalence. Some datasets will have a majority class which is very difficult to distinguish from background voxels but distinctive minority classes, for example disseminated peritoneal metastasis vs lung nodules. In such a case the class prevalence weighting method is clearly ill-suited to the task because there is a mismatch between class “difficulty” and class prevalence. What we actually want to weight the classes based on is “relative difficulty” which is challenging to quantify a priori. Class-prevalence weighting also makes it artificially harder for the network to learn the majority classes. This is disadvantageous as classifying non-majority classes is easier with a well learned majority class due to the reduced possibility space.

\begin{table*}
\caption{The effect of applying loss scheduling on 3D network structures with plain Dice loss. Overall mean Dice improves in each case. Bold indicates improvement over plain loss for individual classes. Star $(*)$ indicates poorly performing classes. }
\label{results_table}
\vskip 0.15in
\begin{center}
\begin{small}
\begin{sc}
\begin{tabular}{lr
|llll|ll|ll}

\bottomrule

        \vphantom{$\int^S_S$}
        Class & Voxels(\%) 
        & UNETR & +\rone\ & +\rtwo\ & + \rthree\
        & 3DUNet & + \rthree\ 
        & Lambda3D & + \rthree\  
        \\

\hline

    \vphantom{$\int^S_S$}
    Mean & 
    & 0.777 & \textbf{0.787} & 0.765 & \textbf{0.792} 
    & 0.484 & \textbf{0.588}
    & 0.707 & \textbf{0.725}
        \\


    
\toprule
\bottomrule

        Liver & 55.2\% 
        & 0.959& 0.959 & 0.939 & 0.952 
        & 0.932 & \textbf{0.945} 
        & 0.914 & \textbf{0.934} 
        \\
        
        Stomach &14.1\% 
        & 0.779* & \textbf{0.814} & 0.748 & \textbf{0.784} 
        & 0.739* & \textbf{0.777} 
        & 0.663* & \textbf{0.678} 
        \\
        
        Spleen &9.1\% 
        & 0.916 & \textbf{0.925} & 0.916 & \textbf{0.917} 
        & 0.831 & 0.788 
        & 0.797* &\textbf{0.932} 
        \\
        
        R kidney &5.1\% 
        & 0.932 & \textbf{0.935} & \textbf{0.935} & 0.932 
        & 0.814 & \textbf{0.920} 
        & 0.811 & \textbf{0.862} 
        \\
        
        L kidney &5.1\% 
        & 0.932 & 0.928 & 0.931 & 0.923
        & 0.623* & \textbf{0.912} 
        & 0.831 & 0.818  
        \\
        
        Aorta &3.1\% 
        & 0.873 & 0.865 & 0.870 & 0.863
        & 0.689* & \textbf{0.692} 
        & 0.840 & \textbf{0.869}  
        \\
        
        IVC &2.8\% 
        & 0.797* & 0.792 & 0.775 & \textbf{0.804} 
        & 0.441* & \textbf{0.711} 
        & 0.751* & \textbf{0.771} 
        \\
        
        Pancreas &2.8\% 
        & 0.738* & 0.735 & 0.698 & \textbf{0.745} 
        & 0.474* & 0.354 
        & 0.526* & 0.505 
        \\
        
        Veins &1.1\%  
        & 0.669* & \textbf{0.680} & \textbf{0.688} & \textbf{0.680} 
        & 0.455* & \textbf{0.483} 
        & 0.663* & \textbf{0.667}
        \\
        
        Gallbladder &0.9\% 
        & 0.590* & \textbf{0.617} & 0.547 & \textbf{0.718} 
        & 0.269* & \textbf{0.423} 
        & 0.514* & 0.484 
        \\
        
        Oesophagus &0.5\%  
        & 0.675* & \textbf{0.704} & \textbf{0.713} & \textbf{0.707} 
        & 0.157* & \textbf{0.303} 
        & 0.678* & \textbf{0.756}
        \\
        
        R Adrenal &0.1\% 
        & 0.634* & \textbf{0.661} & 0.631 & \textbf{0.644} 
        & 0.348* & 0 
        & 0.653* & 0.630 
        \\
        
        L Adrenal &0.1\% 
        & 0.615* & \textbf{0.622} & 0.558 & \textbf{0.629}
        & 0* & 0 
        & 0.554* & 0.517  
        \\
\toprule
\end{tabular}
\end{sc}
\end{small}
\end{center}
\vskip -0.1in
\end{table*}

\subsection{Dynamic Weighting of Class Loss}

We propose a loss scheduling method to address the challenge of imbalanced multi-class learning while minimising the above issues. We propose dynamically adjusting the class contributions to the loss function during the learning process in order to optimise overall performance. We define a set of class weights 
$$
\{w_i \}_t,i\in1,..,N, 
~~~w_i\in[0,1],~~~\{w_i \}_0=1, \forall i\in1,..,N
$$
where $N$ is the number of classes and $t$ is the epoch, and $S_i$ are the class performance scores in terms of Dice Loss:
$$
\{S_i \}_t,i\in1,..,N,~~~S_i\in[0,1],~~~\{S_i \}_0=1 \forall i\in1,..,N
$$
at the end of each epoch we re-weight the $\{w_i\}$. This method aims to address the above concerns in the following ways. Firstly it is closer to weighting by `difficulty' because the weighting is based on current performance which is the best metric we have for `difficulty' during training. Secondly, by changing the loss dynamically throughout the process we still allow the network to learn "easy" classes quickly before re-weighting to promote learning for more `difficult' ones. We want to be careful, however, not to re-weight in such a way that the loss function becomes unstable and difficult to optimise. We first propose the update rule \textbf{\rone}:
\begin{equation} 
w_i=
    \begin{cases}
      0.1, & \text{if}\ S_i>0.9 \\
      1, & \text{otherwise}
    \end{cases} 
\end{equation}
\rone, or \textit{Threshold Class Freezing} is designed to `freeze' learning of classes that pass an empirically determined threshold of good performance, here set to $0.9$. This update rule allows for fast learning of “easy” classes and does not promote instability as we reduce the contribution of classes which already contribute little to the overall loss. As a trade-off we hypothesise that any positive impact on the overall performance must be minimal.

In the current form the re-weighting is conditional on class performance reaching $0.9$, which is never guaranteed. A method with a more universally applicable trigger criterion can be defined for current iteration $t=j$ as \textbf{\rtwo}:
\begin{align}
    C &= |\text{max}(\{S_{i_t}\}, t\in[\text{max}(j-\delta, u), j] \nonumber\\
      &-\text{max}(\{S_{i_t}\}, t\in[\text{max}(0,u),j])| \nonumber
\end{align}
$$i=\text{argmax}(\{S_{i}\}/\{S_{i},w_{i}=0.1\})$$
\begin{equation}
    w_{i_{t=j}}=0.1 \: \text{~~~~if} \: C>\epsilon
\end{equation}
Where $u$ is the iteration where the update rule was last triggered. This update rule requires the local maximum of performance within an epoch window, $\delta$, to be within some small $\epsilon$ of the global performance maximum for a given class else the weighting of the class is decreased to $0.1$. \rtwo, or \textit{Plateau Class Freezing} is designed to freeze the learning of well performing classes based on the plateau trigger condition outlined above. 
Tying the weight update to performance stagnation rather than a given value allows for classes which have reached their maximum at any point to be deprioritised in favour of classes which are still improving. We restrict the $i$ to only the best performing class with a weighting over $0.1$ in order to allow the adjusted loss function to have an effect on the remaining classes.

To further develop the method we experiment with a reweighting that is more focused on addressing the poorer performing classes. We propose \textbf{\rthree}, which uses the same plateau trigger condition from $(3)$.
We then define:
$$
k=\{S_k\}/\{S_k,w_k=0.1\}
$$
\begin{equation}
w_{k_{t=j}}=\text{min}(\frac{1}{S_{i_{t=j}}+\zeta}, 2) \: \text{~~~if} \: C>\epsilon
\end{equation}
With some small $\zeta$ to ensure our values are defined. \rthree, or \textit{Class Boost Strategy} is designed to boost the class weighting based on relative performance of each class, with the trigger condition being the same as in \rtwo for more general application.
This variation on the update rule relates the weighting much more closely to the relative performance for all the classes which have not plateaued. This promotes a learning path which improves performance on these classes but also keeps the range of weights within $[0,2]$ to ensure we do not end up with an unstable loss function.


\section{Results and Discussion}

Our experiments investigate when the strategies help improve the segmentation of challenging or poorly-performing classes, and the impact that these strategies have on the segmentation of better-performing classes. A class is considered to be performing poorly if its Dice metric scores under 0.8. 
We report two sets of experiments:
one where each strategy is applied to UNETR;
another where \rthree\ is applied to different network structures (3DUNet, Lambda3D, UNETR).
Table~\ref{results_table} evaluates performance in all cases using the Dice metric. This is not necessarily the most appropriate metric in segmentation tasks but is universally used, and consistent with the loss function.

\subsection{Scheduling strategy comparison}

This experiment compares UNETR in its original form with training in the presence of one of the loss scheduling methods \rone, \rtwo, \rthree. 
The network's attention is focused dynamically on poorer performing classes. 
Performance on the veins was improved by each of the strategies. Every other poorly performing class gets segmented more effectively by at least one strategy. 

Of the three, \rthree\ fulfils the original intent of improving on {\em all\/} the poorly performing classes. 
Some classes which originally performed well have also been improved (spleen), whereas some have decreased in performance, albeit only slightly (liver, kidneys).

\rtwo\ was the worst-performing, generally decreasing the performance of the classes, this is potentially due to the plateau freezing destabilizing training. Freezing of poorer performing classes disrupts the loss far more than freezing the better performing classes, which may be making learning difficult. \rone\ performed well for a simple method, serving as a good indicator that the hypothesis is correct and that the dynamic attention is able to improve performance of more difficult classes even with a very simple weight update rule.

\subsection{Scheduling strategy generalisability}

This experiment evaluates the effect of \rthree, the most successful of the three scheduling strategies, on three different 3D segmentation networks: UNETR, 3DUNet and Lambda3D (\S\ref{sec-nets}). Every network's  overall mean Dice score is improved. 
This effect is shown most dramatically in 3DUNet, which improved in almost every class when the loss scheduling was included and improved dramatically overall. 
This illustrates that dynamic multi-class scheduling can be applied to any model architecture. It is also not dependent on the choice of loss function (omitted here).



\section{Conclusions and Future Work}

Prevalent classes (stomach) are shown to perform poorly. 
Class prevalence must never be assumed as an indicator of {\em class difficulty}, which is the quality that we are in reality concerned with estimating. 
The proposed strategies focus the attention of a network towards poorer performing classes, balancing the segmentation performance per-class.
They are generic, and can be implemented irrespective of the choice of model or loss function, allowing for this method to be applied to a wide range of problems very easily.
The most sophisticated, \rthree, brings an overall improvement to  all poorly performing classes.


Our Lambda3D extension of the Lambda network is a viable 3D segmentation network. It outperforms the best method (UNETR) on the spleen and oesophagus. It is outperformed on other classes but we plan to further develop the model with other strategies such as varying patch size and domain-knowledge-inspired priors on the attention function.

The loss strategies improve segmentation where the baseline performs averagely; however, models which fail to identify any voxels (adrenal glands) do not recover from failing to generate a good enough embedding. This finding requires further investigation but, intuitively, it is hard to draw attention to an absent feature. The \rthree\ weight update rule has significant scope for further iteration; the effect of cumulative weighting should be explored further; as should the use of evaluation metrics other than Dice.


\section{Compliance with Ethical Standards}
This study was conducted retrospectively using publicly available human subject data. 
The authors acknowledge the use of the University of Oxford Advanced Research Computing (ARC) facility
http://dx.doi.org/10.5281/zenodo.22558.

\bibliography{3Dseg-abridged}
\bibliographystyle{IEEEbib}

\end{document}